\documentclass[10pt,twocolumn,letterpaper]{article}

\usepackage{multirow}
\usepackage[pagenumbers]{iccv} 

\definecolor{iccvblue}{rgb}{0.21,0.49,0.74}
\usepackage[pagebackref,breaklinks,colorlinks,allcolors=iccvblue]{hyperref}

\def\paperID{6018} 
\def\confName{ICCV}
\def\confYear{2025}

\title{Cross-Domain Underwater Image Enhancement Guided by No-Reference Image Quality Assessment: A Transfer Learning Approach}

\author{
Zhi Zhang\\
\and
Minfu Li\\
\and
Lu Li\\
\and
Daoyi Chen\\
*Corresponding author
}

\begin{document}
\maketitle
\begin{abstract}
Single underwater image enhancement (UIE) is a challenging ill-posed problem, but its development is hindered by two major issues: (1) The labels in underwater reference datasets are pseudo labels, relying on these pseudo ground truths in supervised learning leads to domain discrepancy. (2) Underwater reference datasets are scarce, making training on such small datasets prone to overfitting and distribution shift. To address these challenges, we propose Trans-UIE, a transfer learning-based UIE model that captures the fundamental paradigms of UIE through pretraining and utilizes a dataset composed of both reference and non-reference datasets for fine-tuning. However, fine-tuning the model using only reconstruction loss may introduce confirmation bias. To mitigate this, our method leverages no-reference image quality assessment (NR-IQA) metrics from above-water scenes to guide the transfer learning process across domains while generating enhanced images with the style of the above-water image domain. Additionally, to reduce the risk of overfitting during the pretraining stage, we introduce Pearson correlation loss. Experimental results on both full-reference and no-reference underwater benchmark datasets demonstrate that Trans-UIE significantly outperforms state-of-the-art methods.
\end{abstract}
    
\section{Introduction}
\label{sec:intro}
Underwater environments often challenge image quality due to light scattering and inconsistent illumination. The presence of suspended particles further exacerbates issues, leading to common problems like blurring and color distortion in underwater images \cite{article01,article02}. These factors contribute to the degraded visual clarity typically observed in underwater scenes. Underwater image enhancement (UIE) \cite{article03,article04} is a critical technique for acquiring underwater images and surveying underwater environments. It has wide-ranging applications in fields such as ocean exploration, biology, archaeology, and underwater robotics. Therefore, innovations in UIE are of great significance.

\begin{figure}[t]
    \centering
     
    \begin{subfigure}{0.5\textwidth}
        \centering
        \includegraphics[width=\linewidth]{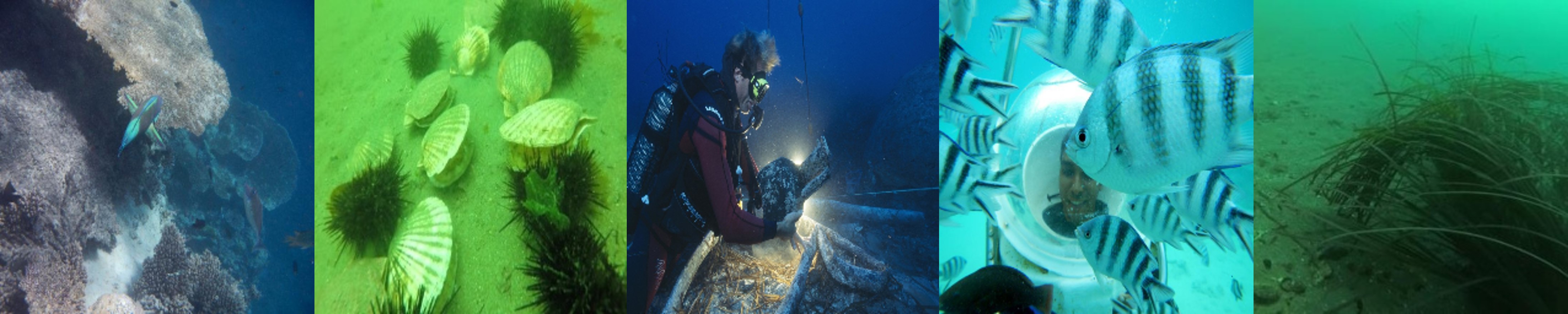} 
        \caption{Input}
        \label{fig:short-a}
    \end{subfigure}%
    
    \begin{subfigure}{0.5\textwidth}
        \centering
        \includegraphics[width=\linewidth]{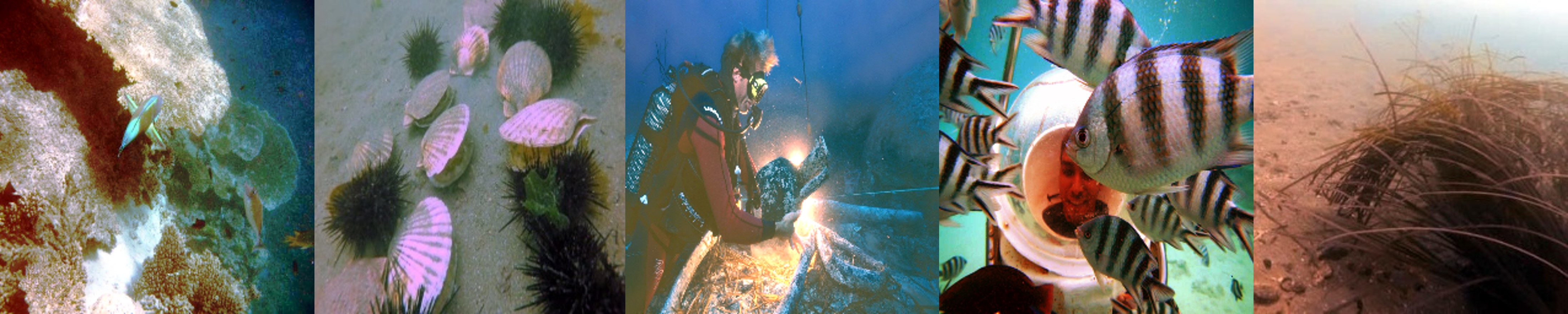} 
        
        \caption{Pseudo Label}
        \label{fig:short-b}
    \end{subfigure}%
    
    \caption{Examples from LSUI \cite{article14} and UIEB \cite{article12} benchmarks. The pseudo labels shown in (b) exhibit issues such as color cast and over-enhancement.}
    \label{fig:one}
\end{figure}

Recently, numerous deep learning-based methods have been proposed to address image restoration problems \cite{article06,article07,article08}. 
Extensive research has also been conducted in the specific area of underwater image restoration \cite{article09,article10,article11,article12,article13}. Compared with traditional methods relying on handcrafted priors, deep learning-based solutions can provide better restoration results due to their data-driven nature.

Despite the success of deep learning methods, most approaches that employ supervised learning to model the distribution of reference datasets still face inherent limitations. First, the labels in underwater reference datasets are pseudo ground truths. These pseudo ground truths are typically generated by processing degraded images with one or more underwater image restoration algorithms and selecting the best output as the pseudo ground truth. For instance, \cite{article44} uses GANs to synthesize underwater images, while LSUI \cite{article14} constructs pseudo labels by processing images with multiple conventional methods and choosing the best result based on evaluation scores. Due to the performance limitations of these algorithms, the quality of pseudo ground truths does not match that of actual above-water images. Over-reliance on pseudo labels during training exacerbates the domain discrepancy between the underwater and above-water image domains. Second, underwater image reference datasets are scarce. For example, the LSUI \cite{article14} dataset contains only 4279 image pairs, while the UIEB \cite{article12} dataset comprises just 890 pairs. Training on such small datasets risks overfitting, leading to distribution shift, which ultimately results in poor performance in real underwater environments. In contrast, non-reference underwater datasets are abundant but cannot be fully utilized. \cref{fig:one} shows examples from the LSUI \cite{article14} and UIEB \cite{article12} benchmarks. The pseudo labels shown exhibit issues such as color cast and over-enhancement. The use of poor pseudo labels limits the performance of reference-trained models.

These issues have significantly hindered progress in underwater image enhancement. To address these challenges, we propose an underwater image enhancement algorithm based on a transfer learning framework, named Trans-UIE. The goal is to reduce reliance on underwater reference datasets, improve the model's generalization ability, and narrow the domain gap between underwater image datasets and real above-water images, further reducing issues such as color cast and insufficient contrast. Specifically, we designed an efficient pre-trained model to learn the fundamental image enhancement paradigm and fine-tuned it using a dataset composed of both reference and non-reference datasets. To prevent confirmation bias caused by relying solely on reconstruction loss during fine-tuning, we guided the fine-tuning process through above-water no-reference image quality assessment (NR-IQA) metrics, enabling cross-domain transfer learning. This approach not only enhances the model's generalization ability on real underwater images but also reduces the domain discrepancy between UIE images and real above-water images. Additionally, to mitigate the risk of overfitting during the pretraining stage, we introduce Pearson correlation loss, allowing the model to focus more on the linear correlations between datasets.

The main contributions are summarized as follows:
\begin{itemize}
    \item We propose Trans-UIE, an underwater image enhancement model based on a transfer learning framework. By fine-tuning a pre-trained model using a dataset composed of both reference and non-reference dataset the optimized model can more accurately restore underwater images.
    
    \item We designed a cross-domain fine-tuning method guided by no-reference image quality assessment (NR-IQA) metrics. This strategy effectively enhances the model's generalization ability on real underwater images and reduces the domain gap between enhanced images and real above-water scenes.
    
    \item We develop a efficient pre-trained model to learn the core paradigm of underwater image enhancement and incorporate Pearson correlation loss to mitigate the risk of overfitting.
    
    \item Extensive experimental results demonstrate the effectiveness of our method.
\end{itemize}


\begin{figure*}
  \centering
  \includegraphics[width=\linewidth]{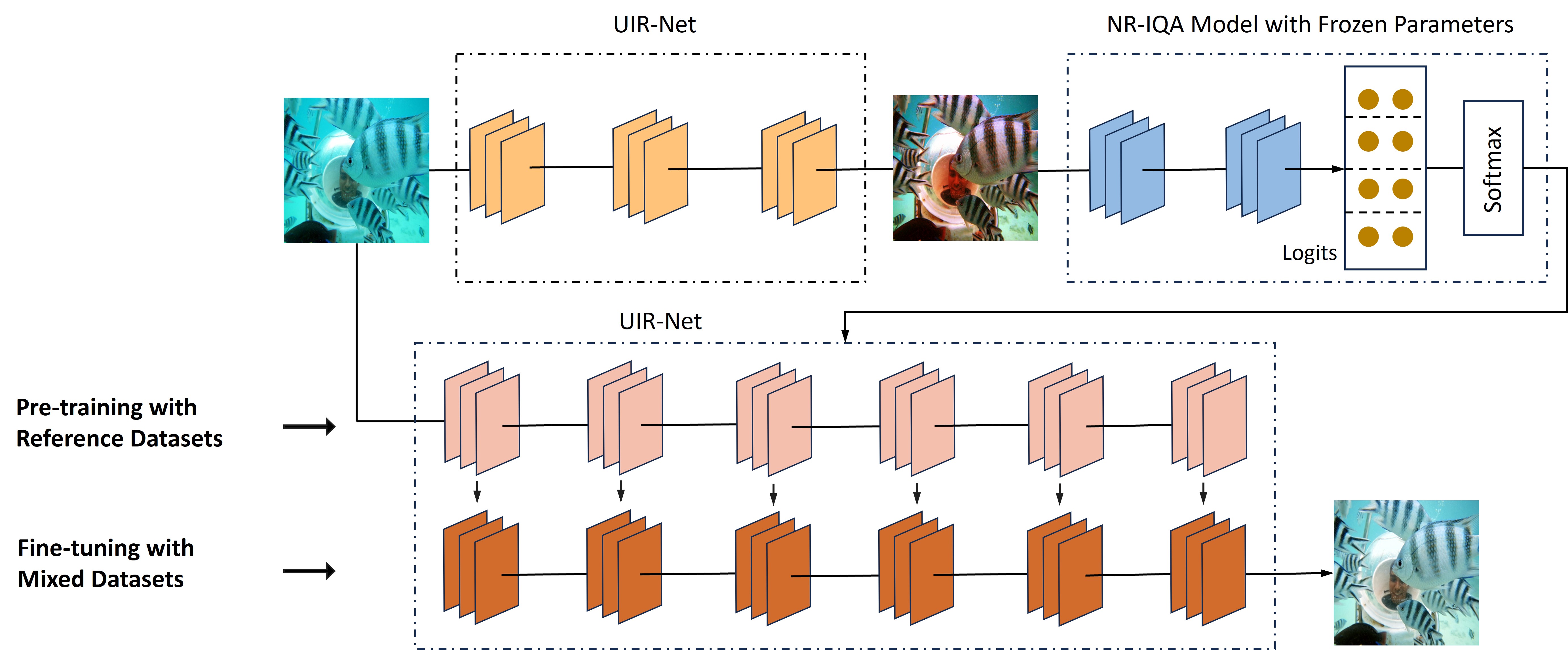} 
  \caption{Illustration of our framework Trans-UIE. Trans-UIE is built upon a transfer learning framework that consists of both pre-training and fine-tuning stages. During pre-training, Trans-UIE learns from reference datasets. To prevent confirmation bias, we incorporate an above-water NR-IQA model with frozen parameters to guide the fine-tuning process. Additionally, a dataset composed of both reference and non-reference datasets is utilized for fine-tuning to mitigate overfitting, reduce distribution shift, and bridge the gap between pseudo labels and the above-water image domain.}
  \label{fig:two}
\end{figure*}

\section{Related Work}
\subsection{Underwater Image Restoration Methods}

Currently, underwater image enhancement methods can be broadly classified into three categories: model-based methods, image enhancement methods, and data-driven methods. The first two categories are considered traditional methods.

Model-based methods (e.g., \cite{article22, article23, article24}) typically rely on handcrafted priors to estimate key parameters in the underwater imaging model, such as transmission and ambient light. While effective in specific scenarios, these prior-based approaches often struggle in more complex real-world environments. Image enhancement methods, such as CLAHE \cite{article25}, Retinex \cite{article26}, fusion methods \cite{article27}, and multi-scale multi-exposure fusion (MMLE) \cite{article28}, aim to improve visual quality by enhancing image details and contrast. Despite their applicability, these methods still face limitations in dynamic lighting conditions.

With the advancement of deep learning, data-driven underwater image enhancement methods have emerged and demonstrated strong competitiveness. In the early stages, due to the lack of paired training samples, researchers introduced generative adversarial networks (GANs) to improve underwater image quality, such as WaterGAN \cite{article29}, FUnIE \cite{article10}, UGAN \cite{article30}, and UIE-DAL \cite{article31}. The release of underwater image datasets such as UIEB \cite{article12} and LSUI \cite{article14} has facilitated the development of more complex network architectures to optimize image enhancement performance, including WaterNet \cite{article12}, Ucolor \cite{article11}, TSDA \cite{article32}, and Ushape \cite{article14}. These models incorporate various innovative techniques to better adapt to complex underwater scenarios. For example, Li et al. \cite{article11} proposed a transmission map-guided network that combines medium transmission maps with multi-channel inputs to enhance model robustness, while PUIE-Net \cite{article33} integrates a conditional variational autoencoder with adaptive instance normalization for adjustable image enhancement. Additionally, Ushape \cite{article14} and RCTNet \cite{article34} introduce Transformer-based architectures and attention mechanisms, significantly improving enhancement performance under complex underwater lighting and color variations.

\subsection{Transfer Learning}

Transfer learning has consistently played a significant role in the field of computer vision \cite{article45, article46}. With the continuous advancement of deep neural networks, transfer learning strategies based on pre-trained models have become a key approach to addressing data scarcity and task migration challenges \cite{article47, article48, article49, article50}. Traditional methods typically train network models on large-scale datasets (e.g., ImageNet) and then fine-tune them on specific tasks or new datasets to achieve knowledge transfer. In existing studies, some approaches advocate fine-tuning the entire network to maximize adaptation to the new domain \cite{article51}, while others modify only the final layers to retain original feature representations \cite{article52}. For example, Kornblith et al. \cite{article54} conducted an in-depth study on different fine-tuning strategies and found that more powerful ImageNet pre-trained models generally achieve better transfer performance on downstream tasks. Yang et al. \cite{article58} proposed using relational graphs as transferable representations instead of relying solely on feature vectors. Moreover, Li et al. \cite{article59} demonstrated that maintaining feature distribution consistency during fine-tuning, along with applying regularization strategies, helps mitigate overfitting and improve model generalization.

To the best of our knowledge, cross-domain transfer learning has not yet been explored in underwater image enhancement. Given the challenges of scarce underwater reference datasets, generalization issues, and the limitations of pseudo ground truths, transfer learning presents a promising solution. This insight motivates the design of Trans-UIE, an underwater image enhancement model based on a transfer learning framework.

\section{Method}
\subsection{Transfer Learning Framework}

Our transfer learning framework is illustrated in \cref{fig:two}. First, a pre-trained model, UIR-Net, is trained on a reference dataset by minimizing the pixel-level loss \( \mathcal{L}_{\text{pix}} \) and the Pearson correlation loss \( \mathcal{L}_{\text{cor}} \) to learn the mapping from degraded images to pseudo labels \( I^* \), thereby obtaining a basic enhancement network. Next, UIR-Net is used to generate pseudo labels for the fine-tuning dataset, providing an initial mapping for the fine-tuning stage. For quality assessment of non-reference images, \cite{article60} indicates that the deep learning-based MUSIQ \cite{article61} is highly reliable; thus, we use MUSIQ \cite{article61} as our above-water NR-IQA to score the generated pseudo labels, obtaining the initial reference quality score \( Q_{\text{reference}} \).

During the fine-tuning stage, we construct a wrapper that integrates UIR-Net with the above-water NR-IQA model. The network parameters of the NR-IQA model are frozen and do not participate in the updating process. Finally, the pre-trained model is fine-tuned by minimizing the following total loss:
\begin{equation}\label{eqn1}
    \begin{split}
        \mathcal{L}_{\text{total}} = \, & \lambda_1 \, \mathcal{L}_{\text{pix}}(I^*, \hat{I}) + \lambda_2 \, \mathcal{L}_{\text{vgg}}(I^*, \hat{I}) \\
        & - \lambda_3 \left( Q(\hat{I}) - Q_{\text{reference}} \right).
    \end{split}
\end{equation}
where \( \lambda_1 \), \( \lambda_2 \), and \( \lambda_3 \) denote the weight coefficients for the pixel-level loss, VGG \cite{article62} perceptual loss, and NR-IQA score loss.  \( Q(\hat{I}) \) is the score assigned to the enhanced image \( \hat{I} \) by NR-IQA within the wrapper. \( Q_{\text{reference}} \) is the reference quality score obtained by evaluating the pseudo labels generated by UIR-Net using NR-IQA, it is responsible for adjusting the learning rate of \(L_{\text{MUSIQ}}\) and does not participate in gradient computation. 


\subsection{Domain Discrepancy and Distribution Shift }

In underwater image enhancement tasks, there is a domain discrepancy between the pseudo labels in the reference dataset and above-water images. Additionally, due to the small size of the reference datasets, supervised learning pre-trained models are prone to overfitting, which leads to distribution shifts with non-reference datasets and adversely affect the model's generalization performance. To address these two issues, we designed the Trans-UIE framework along with an appropriate loss function.  The following will present the detailed mathematical modeling and corresponding formulas.

\noindent \textbf{Domain Discrepancy Caused by Pseudo-label Noise.} Pseudo-labels \(I^*\) are typically generated using conventional methods, and their relationship with the real above-water images \(I_{\text{real}}\) can be expressed as: 
\begin{equation}\label{eq:pseudo_relation_en}
\begin{split}
I^* = I_{\text{real}} + \epsilon_{\text{pseudo}},
\end{split}
\end{equation}
where \(\epsilon_{\text{pseudo}}\) represents the noise introduced by the pseudo-labels. The corresponding domain discrepancy is quantified by: 
\begin{equation}\label{eq:domain_gap_en}
\begin{split}
\Delta_{\text{domain}} = \big[\|I^* - I_{\text{real}}\|_2^2\big].
\end{split}
\end{equation}

Considering the aforementioned aspects, we represent the feature of the pseudo-label \(I^*\) as:
\begin{equation}\label{eq:phi_decompose_en}
\varphi(I^*) = \varphi(I_{\text{real}}) + \Delta_{\varphi}(\epsilon_{\text{pseudo}}).
\end{equation}

We design the overall loss function as \cref{eqn1}. If only the losses \(L_{\text{pix}}\) and \(L_{\text{vgg}}\) are used, the model may learn features that are biased by \(\epsilon_{\text{pseudo}}\), causing the output feature \(\varphi(\hat{I})\) to deviate from the true image feature \(\varphi(I_{\text{real}})\). The introduction of the NR-IQA score loss \(L_{\text{MUSIQ}}\) helps to align the quality of the output images with the target  \(\varphi(I_{\text{real}})\), thereby partially mitigating the domain discrepancy induced by \(\epsilon_{\text{pseudo}}\).

During training, \( L_{\text{total}} \) gradually converges, while \( Q(\hat{I}) \) increases. We assume that the mapping function \( f(\cdot) \) maps the deep features of the image to the NR-IQA score space. When the model output \( \hat{I} \)'s feature approaches the above-water image feature \( \varphi(I_{\text{real}}) \), \( Q(\hat{I}) \) will tend to reach its maximum value. Hence, the total loss can be approximated by:
\begin{equation}\label{eq:total_loss_approx_en}
\begin{split}
L_{\text{total}} \approx \lambda \|\varphi(\hat{I}) - (\varphi(I_{\text{real}}) + \Delta_{\varphi}(\epsilon_{\text{pseudo}}))\|_2^2 \\
\quad + \lambda_3 \|f(\varphi(\hat{I})) - f(\varphi(I_{\text{real}}))\|_2^2, 
\end{split}
\end{equation}
where, the first term ensures that the model output closely recovers the pseudo-label, although the noise \(\Delta_{\varphi}(\epsilon_{\text{pseudo}})\) in the pseudo-label may bias the output features toward the reference domain. The second term, by imposing a constraint in the score space, forces the model output to approach the above-water image features, thereby reducing the bias induced by \(\Delta_{\text{domain}}\). We reduce the impact of the first term by appropriately adjusting the loss weights and effectively achieve the goal.


\noindent \textbf{Feature Distribution Shift.} Assume that an underwater image is a real above-water image corrupted by noise. Specifically, images from the reference and non-reference datasets can be modeled as:
\begin{equation}\label{eq:IR}
I_R = I_{\text{real}} + \epsilon_R,
\end{equation}
\begin{equation}\label{eq:IN}
I_N = I_{\text{real}} + \epsilon_N,
\end{equation}
where $\epsilon_R$ and $\epsilon_N$ denote the noise components in the reference and non-reference images, respectively. Given the near-linearity of the feature extraction network $\varphi(\cdot)$, we have:
\begin{equation}\label{eq:phi_IR}
\varphi(I_R) \approx \varphi(I_{\text{real}}) + \varphi(\epsilon_R),
\end{equation}
\begin{equation}\label{eq:phi_IN}
\varphi(I_N) \approx \varphi(I_{\text{real}}) + \varphi(\epsilon_N).
\end{equation}

Thus, the mean features for the two domains are given by:
\begin{equation}\label{eq:muR}
\mu_R = \mathbb{E}[\varphi(I_R)] \approx \varphi(I_{\text{real}}) + \mathbb{E}[\varphi(\epsilon_R)],
\end{equation}
\begin{equation}\label{eq:muN}
\mu_N = \mathbb{E}[\varphi(I_N)] \approx \varphi(I_{\text{real}}) + \mathbb{E}[\varphi(\epsilon_N)].
\end{equation}

Therefore, the feature distribution shift is defined as:
\begin{equation}\label{eq:featshift}
\Delta_{\text{feat}} = \|\mu_R - \mu_N\|_2^2 \approx \|\mathbb{E}[\varphi(\epsilon_R)] - \mathbb{E}[\varphi(\epsilon_N)]\|_2^2.
\end{equation}

During fine-tuning, we employ a training set that mixes reference and non-reference images and generate pseudo labels using a pre-trained model. The pixel-level loss $L_{\text{pix}}$ and the perceptual loss $L_{\text{vgg}}$ encourage the expected features of the generated image $\hat{I}$ to approach $\mu_R$. Simultaneously, the NR-IQA score loss $L_{\text{MUSIQ}}$ further enhances the quality score $E[Q(\hat{I})]$, driving the model to suppress noise during generation. In this way, the discrepancy between the reference noise $\epsilon_R$ and the non-reference noise $\epsilon_N$ is effectively reduced, thereby compensating for the feature distribution shift $\Delta_{\text{feat}}$.

The overall loss function can be approximated as:
\begin{equation}\label{eq:totalloss}
L_{\text{total}} \approx \lambda \|\varphi(\hat{I}) - \mu_R\|_2^2 + \lambda_3 \left( E[Q(\hat{I})]_{\text{desired}} - E[Q(\hat{I})] \right),
\end{equation}
where $E[Q(\hat{I})]_{\text{desired}}$ denotes the desired average quality score we intend to achieve during fine-tuning, after which training can be stopped. The first term ensures that the features of the generated image remain close to $\mu_R$, preventing the model from deviating, while the second term (weighted by $\lambda_3$) enhances the quality score by reducing noise interference, thereby decreasing the feature distribution shift $\Delta_{\text{feat}}$.


\subsection{Pretrained Network UIR-Net}

\begin{figure}[t]
  \centering
  \includegraphics[scale=0.32]{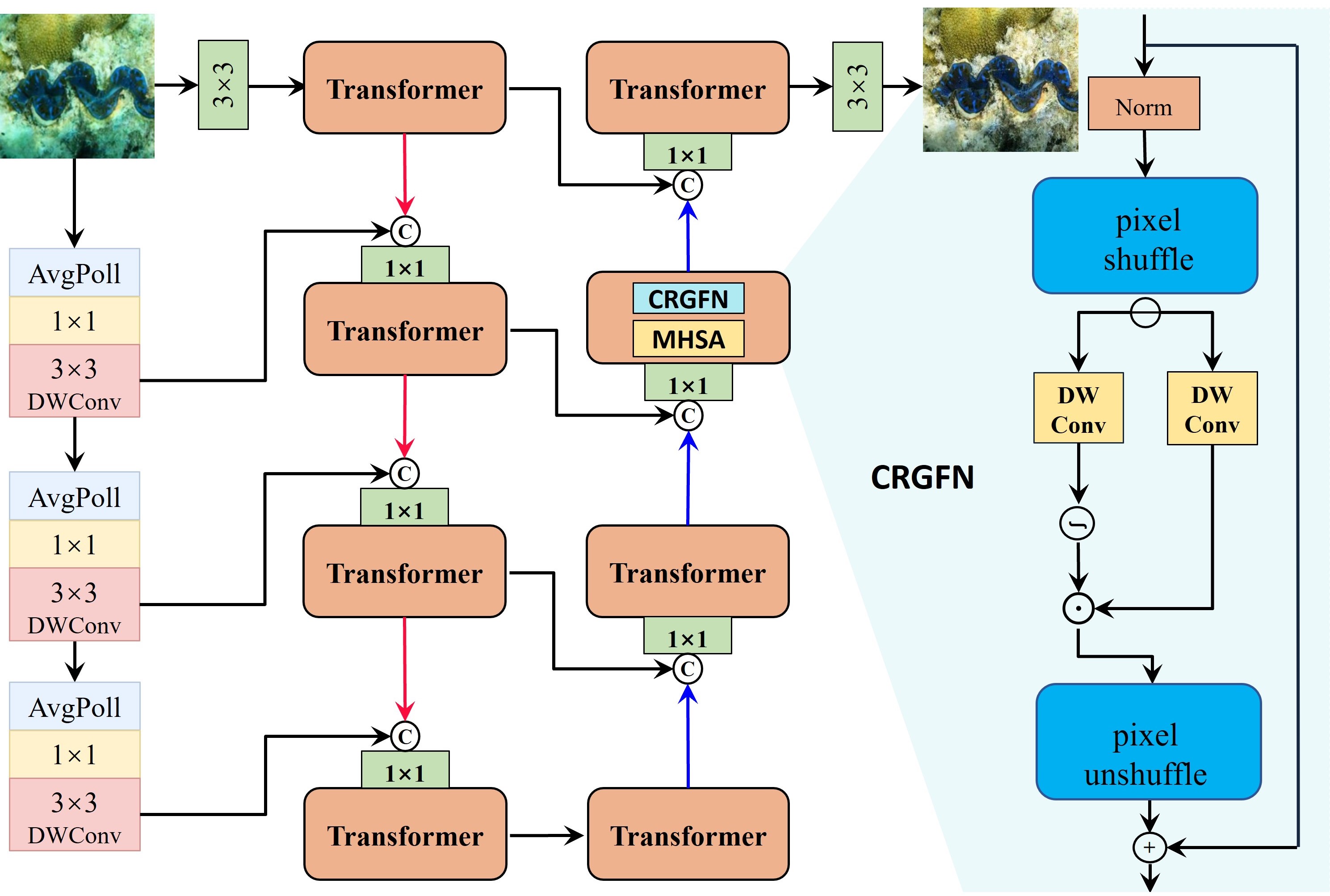} 
  \caption{An overview of the proposed Pre-Train Network (UIR-Net).}
   \label{fig:tree}
\end{figure}

In this study, we propose UIR-Net, a U-shaped Transformer architecture that captures global context and fine-grained details, providing a solid foundation for transfer learning. The framework is shown in \cref{fig:tree}. The input image \( I \in \mathbb{R}^{H \times W \times C} \) is divided into overlapping patches using a \( 3 \times 3 \) convolution. The encoder reduces the spatial resolution through an Unshuffle operation while extracting global features with Transformer modules. The decoder restores the resolution using a Shuffle operation. Skip connections fuse low-level encoder features with high-level decoder semantics.

To enhance the model’s capacity, each Transformer module includes a Channel Reordering Gated Feedforward Network (CRGFN), which reconstructs multi-scale features via channel reordering \cite{article63}, depth-wise convolutions \cite{article76}, and gated fusion \cite{article77}, improving the network’s ability to model degradation factors.

\subsection{Pearson Correlation Loss}

Since the reference dataset is small, multiple rounds of pre-training pose a risk of overfitting, as the model may memorize noise and specific patterns in the data rather than learning the underlying fundamental representations. To mitigate this issue, we incorporate the Pearson correlation \cite{article64} as a loss during the pre-training process to capture the linear relationships between data points. Specifically, for the pseudo-label \(I_{gt}\) and the model output \(\hat{I}\) (or recovered image \(\hat{I}\)), the Pearson correlation coefficient \(\rho(\hat{I}, I_{gt})\) is defined as follows:
\begin{equation}\label{eq:pearson_en}
\begin{split}
\rho(\hat{I}, I_{gt}) = \frac{1}{3HW} \sum_{i=1}^{3HW} \frac{\Bigl(\hat{I}(i) - \mu(\hat{I})\Bigr)\Bigl(I_{gt}(i) - \mu(I_{gt})\Bigr)}
{\sigma(\hat{I})\, \sigma(I_{gt})},
\end{split}
\end{equation}
where \(\hat{I}(i)\) denotes the \(i\)-th pixel of the image, and \(\mu(\hat{I})\) and \(\sigma(\hat{I})\) represent the mean and standard deviation of \(\hat{I}\) (similarly for \(I_{gt}\)). The value of \(\rho(\hat{I}, I_{gt})\) lies in the range \([-1, 1]\); when the two images are perfectly positively correlated, \(\rho(\hat{I}, I_{gt}) = 1\), and when they are negatively correlated, \(\rho(\hat{I}, I_{gt}) = -1\).

Accordingly, the correlation loss is formulated as: 
\begin{equation}\label{eq:cor_loss_en}
\begin{split}
L_{cor} = \frac{1}{2}\Bigl( 1 - \rho(\hat{I}, I_{gt}) \Bigr),
\end{split}
\end{equation}
so that \(L_{cor} = 0\) when the recovered image perfectly aligns with the pseudo-label. By incorporating the Pearson correlation loss, we effectively reduce the risk of overfitting in the pre-training process.

\begin{table}
\resizebox{\linewidth}{!}{
\begin{tabular}{c|cccccc}
\hline
  & \multicolumn{2}{c}{LSUI-400}   & \multicolumn{2}{c}{UIEB-90}    \\
Method & PSNR($\uparrow$) & SSIM($\uparrow$) & PSNR($\uparrow$) & SSIM($\uparrow$)  \\
\hline
GDCP \cite{article72} & 14.46 & 0.7069 & 13.91 & 0.7265  \\
Water-Net \cite{article12} & 22.74 & 0.8560 & 21.04 & 0.8604  \\
Ucolor \cite{article11} & 24.03 & 0.8865 & 20.89 & 0.8732 \\
PUIE-Net \cite{article33} & 24.14 & 0.8762 & 22.98 & 0.8984 \\
Semi-UIR \cite{article60} & \underline{25.83} & \underline{0.9103} & \textbf{23.89} & \underline{0.9008} \\
U-shape \cite{article14} & 24.19 & 0.8793 & 21.56 & 0.8874 \\
HUPE \cite{article73} & 22.79 & 0.8697 & 21.99 & 0.8665 \\
Trans-UIE(ours) & \textbf{27.23} & \textbf{0.9283} & \underline{23.16} & \textbf{0.9013}  \\
\hline
\end{tabular}
}
\caption{Quantitative comparison on the LSUI-400 \cite{article14} and UIEB-90 \cite{article12} datasets. Best results are in \textbf{bold} and the second best results are with \underline{underline}.}
\label{tab:1}
\end{table}

\begin{figure*}
    \centering
    \begin{subfigure}{0.09\linewidth}
        \centering
        \textbf{Input} 
    \end{subfigure}%
    \hfill
    \begin{subfigure}{0.09\linewidth}
        \centering
        \textbf{GDCP} 
    \end{subfigure}%
    \hfill
    \begin{subfigure}{0.09\linewidth}
        \centering
        \textbf{Water-Net} 
    \end{subfigure}%
    \hfill
    \begin{subfigure}{0.09\linewidth}
        \centering
        \textbf{Ucolor} 
    \end{subfigure}%
    \hfill
    \begin{subfigure}{0.09\linewidth}
        \centering
        \textbf{PUIE-Net} 
    \end{subfigure}%
    \hfill
    \begin{subfigure}{0.09\linewidth}
        \centering
        \textbf{Semi-UIR} 
    \end{subfigure}%
    \hfill
    \begin{subfigure}{0.09\linewidth}
        \centering
        \textbf{U-shape} 
    \end{subfigure}%
    \hfill
    \begin{subfigure}{0.09\linewidth}
        \centering
        \textbf{HUPE} 
    \end{subfigure}%
    \hfill
    \begin{subfigure}{0.09\linewidth}
        \centering
        \textbf{Ours} 
    \end{subfigure}%
    \hfill
    \begin{subfigure}{0.09\linewidth}
        \centering
        \textbf{GT} 
    \end{subfigure}%

    \vskip 0.5em
    \begin{subfigure}{0.09\linewidth}
        \centering
        \includegraphics[width=\linewidth]{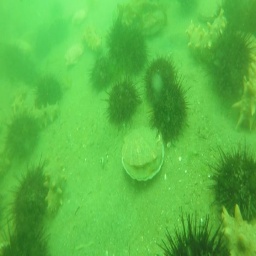} 
    \end{subfigure}%
    \hfill
    \begin{subfigure}{0.09\linewidth}
        \centering
        \includegraphics[width=\linewidth]{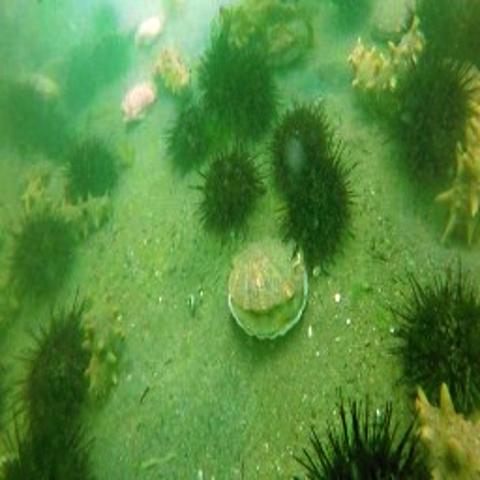} 
    \end{subfigure}%
    \hfill
    \begin{subfigure}{0.09\linewidth}
        \centering
        \includegraphics[width=\linewidth]{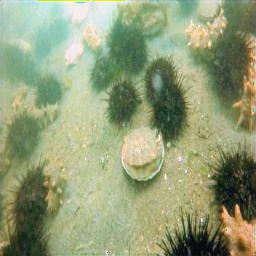} 
    \end{subfigure}%
    \hfill
    \begin{subfigure}{0.09\linewidth}
        \centering
        \includegraphics[width=\linewidth]{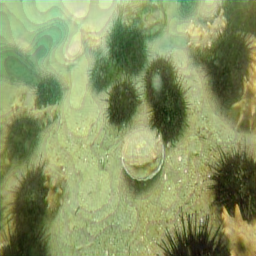} 
    \end{subfigure}%
    \hfill
    \begin{subfigure}{0.09\linewidth}
        \centering
        \includegraphics[width=\linewidth]{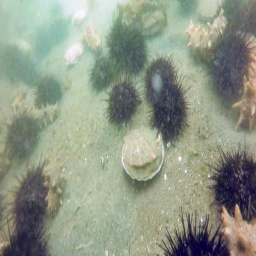} 
    \end{subfigure}%
    \hfill
    \begin{subfigure}{0.09\linewidth}
        \centering
        \includegraphics[width=\linewidth]{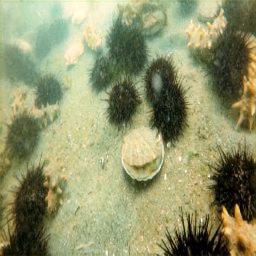} 
    \end{subfigure}%
    \hfill
    \begin{subfigure}{0.09\linewidth}
        \centering
        \includegraphics[width=\linewidth]{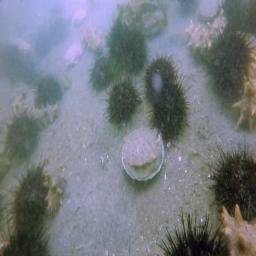} 
    \end{subfigure}%
    \hfill
    \begin{subfigure}{0.09\linewidth}
        \centering
        \includegraphics[width=\linewidth]{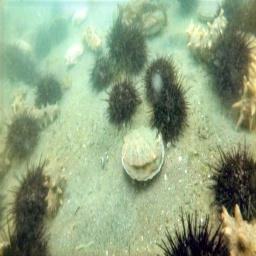} 
    \end{subfigure}%
    \hfill
    \begin{subfigure}{0.09\linewidth}
        \centering
        \includegraphics[width=\linewidth]{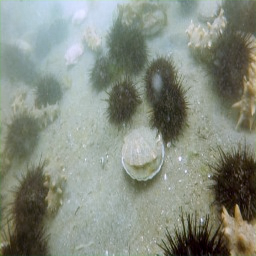} 
    \end{subfigure}%
    \hfill
    \begin{subfigure}{0.09\linewidth}
        \centering
        \includegraphics[width=\linewidth]{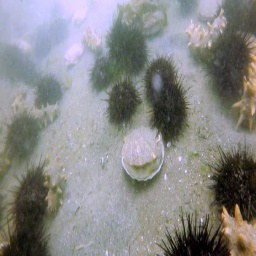} 
    \end{subfigure}%
    \vskip 0.5em
    \begin{subfigure}{0.09\linewidth}
        \centering
        \includegraphics[width=\linewidth]{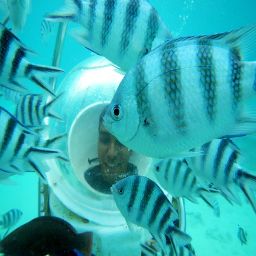} 
    \end{subfigure}%
    \hfill
    \begin{subfigure}{0.09\linewidth}
        \centering
        \includegraphics[width=\linewidth]{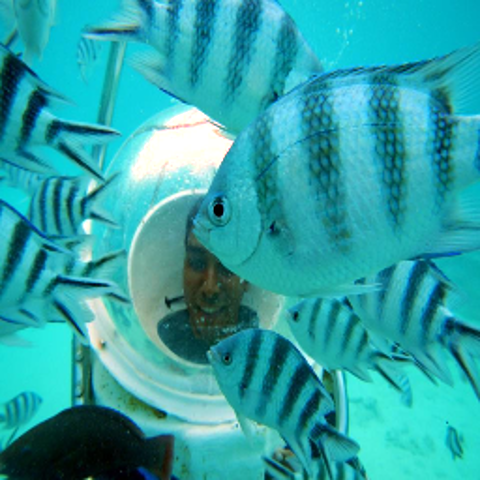} 
    \end{subfigure}%
    \hfill
    \begin{subfigure}{0.09\linewidth}
        \centering
        \includegraphics[width=\linewidth]{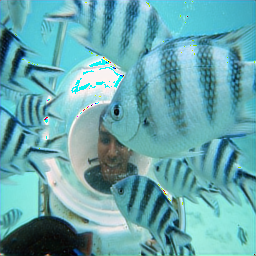} 
    \end{subfigure}%
    \hfill
    \begin{subfigure}{0.09\linewidth}
        \centering
        \includegraphics[width=\linewidth]{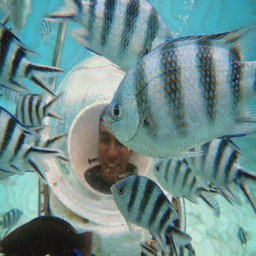} 
    \end{subfigure}%
    \hfill
    \begin{subfigure}{0.09\linewidth}
        \centering
        \includegraphics[width=\linewidth]{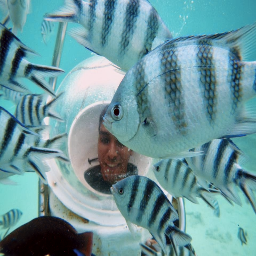} 
    \end{subfigure}%
    \hfill
    \begin{subfigure}{0.09\linewidth}
        \centering
        \includegraphics[width=\linewidth]{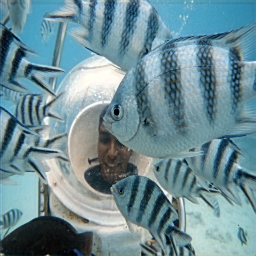} 
    \end{subfigure}%
    \hfill
    \begin{subfigure}{0.09\linewidth}
        \centering
        \includegraphics[width=\linewidth]{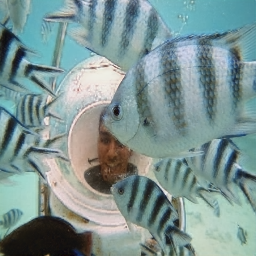} 
    \end{subfigure}%
    \hfill
    \begin{subfigure}{0.09\linewidth}
        \centering
        \includegraphics[width=\linewidth]{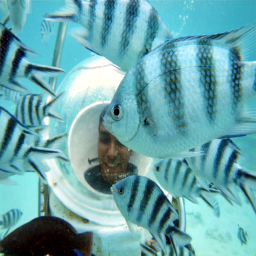} 
    \end{subfigure}%
    \hfill
    \begin{subfigure}{0.09\linewidth}
        \centering
        \includegraphics[width=\linewidth]{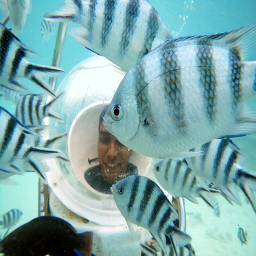} 
    \end{subfigure}%
    \hfill
    \begin{subfigure}{0.09\linewidth}
        \centering
        \includegraphics[width=\linewidth]{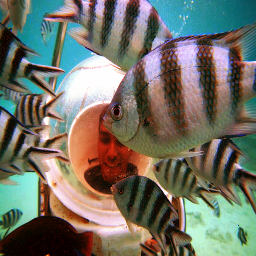} 
    \end{subfigure}%
    \vskip 0.5em
    \begin{subfigure}{0.09\linewidth}
        \centering
        \includegraphics[width=\linewidth]{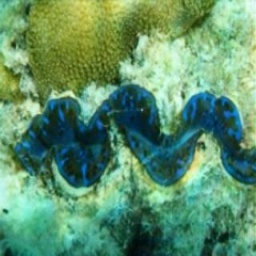} 
    \end{subfigure}%
    \hfill
    \begin{subfigure}{0.09\linewidth}
        \centering
        \includegraphics[width=\linewidth]{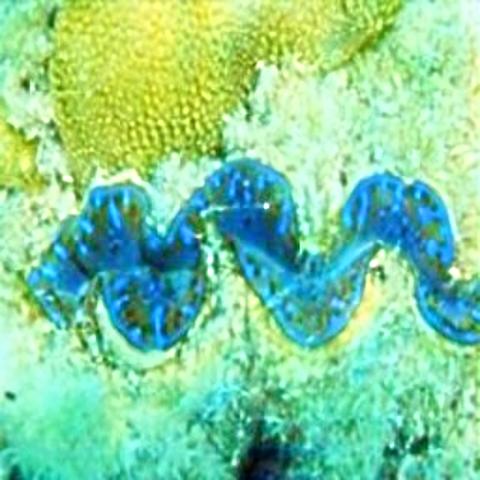} 
    \end{subfigure}%
    \hfill
    \begin{subfigure}{0.09\linewidth}
        \centering
        \includegraphics[width=\linewidth]{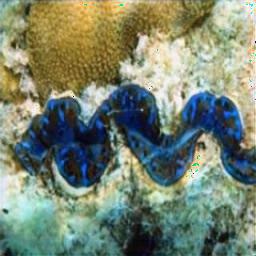} 
    \end{subfigure}%
    \hfill
    \begin{subfigure}{0.09\linewidth}
        \centering
        \includegraphics[width=\linewidth]{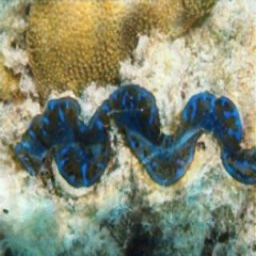} 
    \end{subfigure}%
    \hfill
    \begin{subfigure}{0.09\linewidth}
        \centering
        \includegraphics[width=\linewidth]{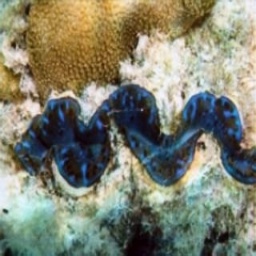} 
    \end{subfigure}%
    \hfill
    \begin{subfigure}{0.09\linewidth}
        \centering
        \includegraphics[width=\linewidth]{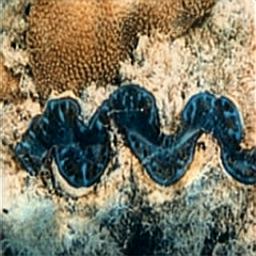} 
    \end{subfigure}%
    \hfill
    \begin{subfigure}{0.09\linewidth}
        \centering
        \includegraphics[width=\linewidth]{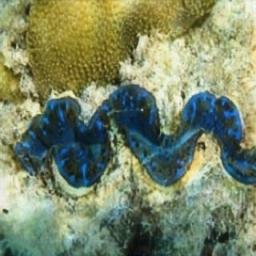} 
    \end{subfigure}%
    \hfill
    \begin{subfigure}{0.09\linewidth}
        \centering
        \includegraphics[width=\linewidth]{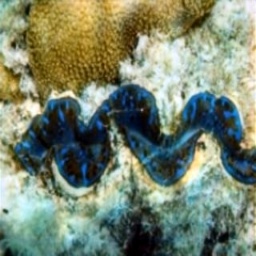} 
    \end{subfigure}%
    \hfill
    \begin{subfigure}{0.09\linewidth}
        \centering
        \includegraphics[width=\linewidth]{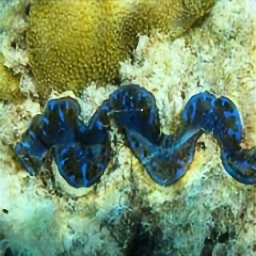} 
    \end{subfigure}%
    \hfill
    \begin{subfigure}{0.09\linewidth}
        \centering
        \includegraphics[width=\linewidth]{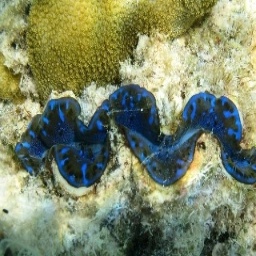} 
    \end{subfigure}%
    \vskip 0.5em
    \begin{subfigure}{0.09\linewidth}
        \centering
        \includegraphics[width=\linewidth]{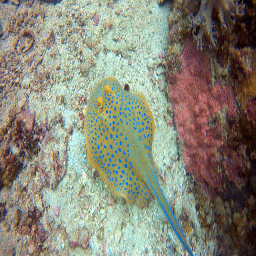} 
    \end{subfigure}%
    \hfill
    \begin{subfigure}{0.09\linewidth}
        \centering
        \includegraphics[width=\linewidth]{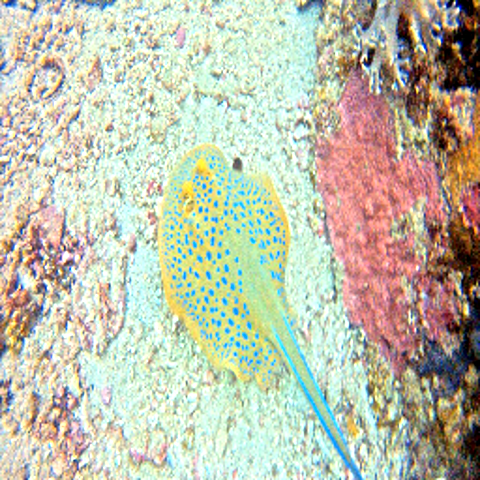} 
    \end{subfigure}%
    \hfill
    \begin{subfigure}{0.09\linewidth}
        \centering
        \includegraphics[width=\linewidth]{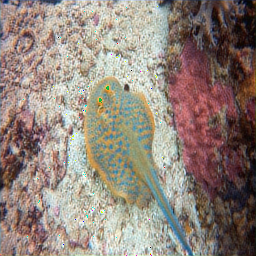} 
    \end{subfigure}%
    \hfill
    \begin{subfigure}{0.09\linewidth}
        \centering
        \includegraphics[width=\linewidth]{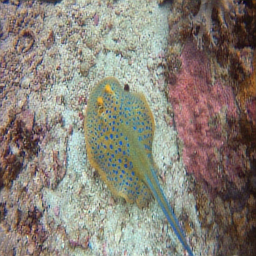} 
    \end{subfigure}%
    \hfill
    \begin{subfigure}{0.09\linewidth}
        \centering
        \includegraphics[width=\linewidth]{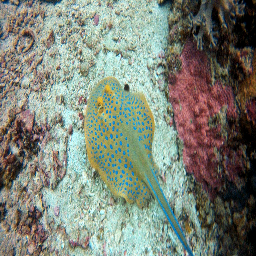} 
    \end{subfigure}%
    \hfill
    \begin{subfigure}{0.09\linewidth}
        \centering
        \includegraphics[width=\linewidth]{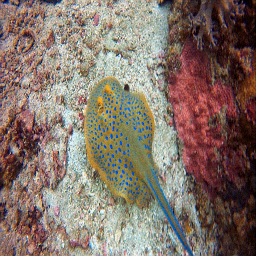} 
    \end{subfigure}%
    \hfill
    \begin{subfigure}{0.09\linewidth}
        \centering
        \includegraphics[width=\linewidth]{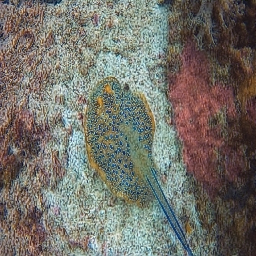} 
    \end{subfigure}%
    \hfill
    \begin{subfigure}{0.09\linewidth}
        \centering
        \includegraphics[width=\linewidth]{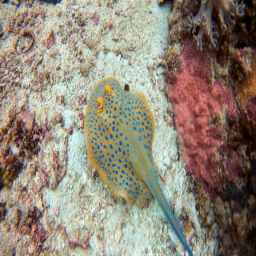} 
    \end{subfigure}%
    \hfill
    \begin{subfigure}{0.09\linewidth}
        \centering
        \includegraphics[width=\linewidth]{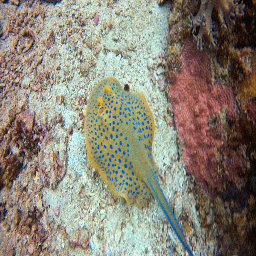} 
    \end{subfigure}%
    \hfill
    \begin{subfigure}{0.09\linewidth}
        \centering
        \includegraphics[width=\linewidth]{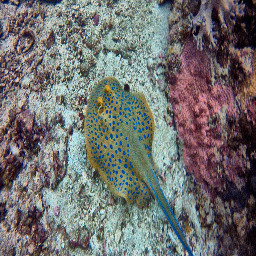} 
    \end{subfigure}%
    
    \caption{Visual comparisons of full-reference data from LSUI-400 \cite{article14} and UIEB-90 \cite{article12} benchmarks. Besides, the ground truths are displayed in the last columns.}
    \label{fig:four}
\end{figure*}
\begin{table*}[ht]
\centering
\resizebox{\textwidth}{!}{
\begin{tabular}{l|ccc|ccc|ccc}
\hline
\multirow{2}{*}{Method} 
& \multicolumn{3}{c|}{UIQM ($\uparrow$)} 
& \multicolumn{3}{c|}{UCIQE ($\uparrow$)} 
& \multicolumn{3}{c}{MUSIQ ($\uparrow$)} \\
\cline{2-10}
& UIEB-60 & EUVP & RUIE 
& UIEB-60 & EUVP & RUIE 
& UIEB-60 & EUVP & RUIE  \\
\hline
GDCP \cite{article72} 
& 2.317 & 2.714 & 2.875 
& 0.5786 & \underline{0.6000} & 0.5616 
& 42.74 & 35.80 & 32.48 \\

Water-Net \cite{article12} 
& \underline{2.857} & \underline{3.058} & \textbf{3.217} 
& \underline{0.5813} & \textbf{0.6071} & \textbf{0.5988} 
& 44.29 & 37.59 & 33.32 \\

Ucolor \cite{article11} 
& \textbf{3.013} & \textbf{3.112} & \underline{3.123} 
& 0.5483 & 0.5705 & 0.5395 
& 43.05 & 39.16 & 34.33 \\

PUIE-Net \cite{article33} 
& 2.688 & 3.047 & 3.053 
& 0.5643 & 0.5757 & 0.5548 
& 46.43 & 40.79 & 32.47 \\

Semi-UIR \cite{article60} 
& 2.759 & 3.047 & 2.949 
& \textbf{0.5855} & 0.5927 & \underline{0.5711} 
& \underline{48.29} & \underline{44.20} & \underline{34.66} \\

U-shape \cite{article14} 
& 2.746 & 2.902 & 2.989 
& 0.5438 & 0.5609 & 0.5552 
& 44.45 & 39.18 & 32.75 \\

HUPE \cite{article73} 
& 2.711 & 3.031 & 2.974 
& 0.5704 & 0.5800 & 0.5544 
& 40.49 & 35.26 & 32.06 \\

Trans-UIE (ours)   
& 2.454 & 2.851 & 2.883 
& 0.5493 & 0.5548 & 0.5617 
& \textbf{49.41} & \textbf{49.62} & \textbf{40.23} \\
\hline
\end{tabular}
}
\caption{Quantitative results on non-reference datasets. Best results are in \textbf{bold}, and the second-best results are underlined.}
\label{tab:2}
\end{table*}

\section{Experiments}

\subsection{Datasets and Evaluation Metrics}
\textbf{Dataset.} During the pre-training phase, our training set consists of 4679 labeled image pairs, with 3879 pairs from the reference dataset LSUI \cite{article14} and 800 pairs from UIEB \cite{article12}. The test set includes the remaining 400 pairs from LSUI \cite{article14} and 90 pairs from UIEB \cite{article12}. In the fine-tuning phase, the training set consists of 10,000 samples, each containing a pseudo-label generated by the pre-trained model and a reference score produced by NR-IQA. Of these, 3830 samples are drawn from the non-reference dataset RUIE \cite{article65}, 3870 from EUVP \cite{article44}, and 2300 from LSUI \cite{article14}. The test set consists of 60 challenging unlabeled images from UIEB-60 \cite{article12}, 400 images from RUIE \cite{article65}, and 330 images from EUVP \cite{article44}. These datasets cover diverse underwater scenes, water types, and lighting conditions.

\noindent \textbf{Evaluation Metrics.} For the reference test set, we use Peak Signal-to-Noise Ratio (PSNR) and Structural Similarity Index (SSIM)  as full-reference metrics to assess image quality. For the non-reference test set, evaluation is conducted using UIQM \cite{article66}, UCIQE \cite{article67} and MUSIQ \cite{article61}. After generating the enhanced images, their sizes are resized to \(256 \times 256\) before applying these metrics for evaluation.

\subsection{Implementation Details}

Our implementation is based on the PyTorch \cite{article69} framework and was trained on an NVIDIA RTX 4090 GPU. During pre-training, random horizontal and vertical flips were applied for data augmentation to increase diversity. The network was trained for approximately 380 epochs using a progressive learning pipeline \cite{article70}, where the batch size was gradually reduced and the input patch size progressively increased. We employed the AdamW \cite{article71} optimizer with a cosine annealing learning rate scheduler \cite{article78}. In pre-training, pixel loss and Pearson correlation loss were used as loss functions, with an initial learning rate of \(3 \times 10^{-4}\). The pre-training process took about 18 hours. In the fine-tuning phase, random flips were similarly applied for data preprocessing. The network was fine-tuned with approximately 1000 optimization steps using a batch size of 2. The AdamW \cite{article71} optimizer was used along with a cosine annealing learning rate scheduler \cite{article78}. The total loss during fine-tuning was a weighted sum of pixel loss, VGG \cite{article62} perceptual loss, and NR-IQA score loss, with an initial learning rate of \(1 \times 10^{-5}\).


\begin{figure*}
    \centering
    \begin{subfigure}{0.09\linewidth}
        \centering
        \textbf{Input} 
    \end{subfigure}%
    \hfill
    \begin{subfigure}{0.09\linewidth}
        \centering
        \textbf{GDCP} 
    \end{subfigure}%
    \hfill
    \begin{subfigure}{0.09\linewidth}
        \centering
        \textbf{Water-Net} 
    \end{subfigure}%
    \hfill
    \begin{subfigure}{0.09\linewidth}
        \centering
        \textbf{Ucolor} 
    \end{subfigure}%
    \hfill
    \begin{subfigure}{0.09\linewidth}
        \centering
        \textbf{PUIE-Net} 
    \end{subfigure}%
    \hfill
    \begin{subfigure}{0.09\linewidth}
        \centering
        \textbf{Semi-UIR} 
    \end{subfigure}%
    \hfill
    \begin{subfigure}{0.09\linewidth}
        \centering
        \textbf{U-shape} 
    \end{subfigure}%
    \hfill
    \begin{subfigure}{0.09\linewidth}
        \centering
        \textbf{HUPE} 
    \end{subfigure}%
    \hfill
    \begin{subfigure}{0.09\linewidth}
        \centering
        \textbf{Ours} 
    \end{subfigure}%

    \vskip 0.5em
    
    \begin{subfigure}{0.09\linewidth}
        \centering
        \includegraphics[width=\linewidth]{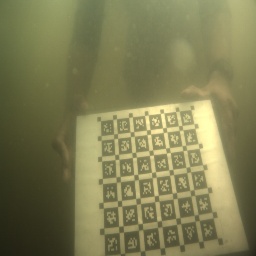} 
    \end{subfigure}%
    \hfill
    \begin{subfigure}{0.09\linewidth}
        \centering
        \includegraphics[width=\linewidth]{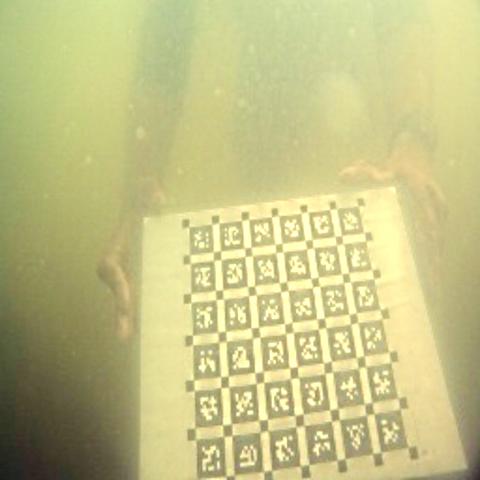} 
    \end{subfigure}%
    \hfill
    \begin{subfigure}{0.09\linewidth}
        \centering
        \includegraphics[width=\linewidth]{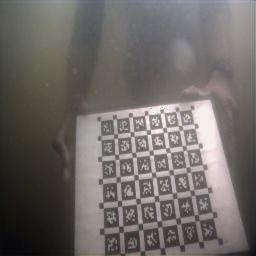} 
    \end{subfigure}%
    \hfill
    \begin{subfigure}{0.09\linewidth}
        \centering
        \includegraphics[width=\linewidth]{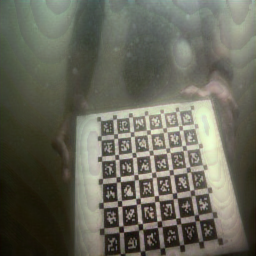} 
    \end{subfigure}%
    \hfill
    \begin{subfigure}{0.09\linewidth}
        \centering
        \includegraphics[width=\linewidth]{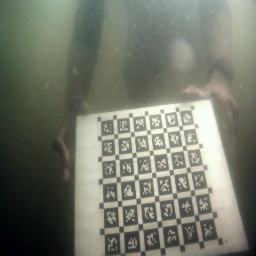} 
    \end{subfigure}%
    \hfill
    \begin{subfigure}{0.09\linewidth}
        \centering
        \includegraphics[width=\linewidth]{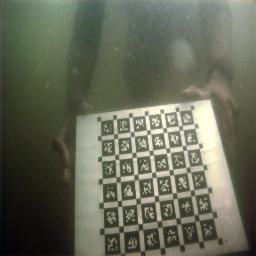} 
    \end{subfigure}%
    \hfill
    \begin{subfigure}{0.09\linewidth}
        \centering
        \includegraphics[width=\linewidth]{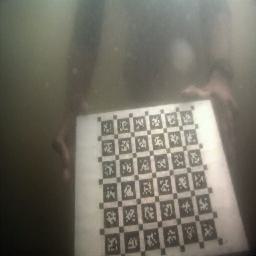} 
    \end{subfigure}%
    \hfill
    \begin{subfigure}{0.09\linewidth}
        \centering
        \includegraphics[width=\linewidth]{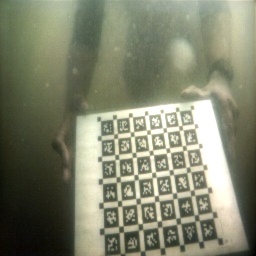} 
    \end{subfigure}%
    \hfill
    \begin{subfigure}{0.09\linewidth}
        \centering
        \includegraphics[width=\linewidth]{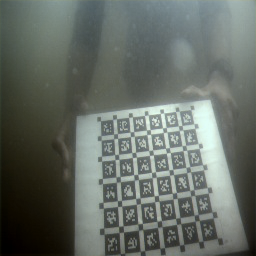} 
    \end{subfigure}%
    \vskip 0.5em
    
    \begin{subfigure}{0.09\linewidth}
        \centering
        \includegraphics[width=\linewidth]{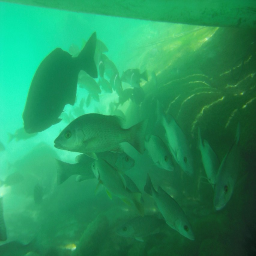} 
    \end{subfigure}%
    \hfill
    \begin{subfigure}{0.09\linewidth}
        \centering
        \includegraphics[width=\linewidth]{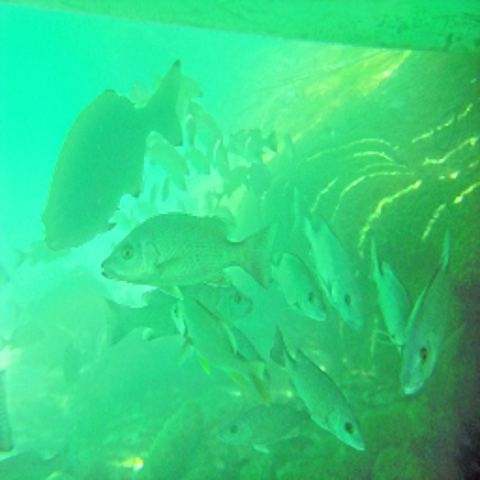} 
    \end{subfigure}%
    \hfill
    \begin{subfigure}{0.09\linewidth}
        \centering
        \includegraphics[width=\linewidth]{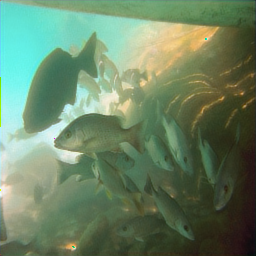} 
    \end{subfigure}%
    \hfill
    \begin{subfigure}{0.09\linewidth}
        \centering
        \includegraphics[width=\linewidth]{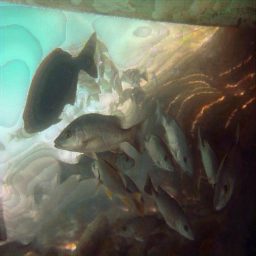} 
    \end{subfigure}%
    \hfill
    \begin{subfigure}{0.09\linewidth}
        \centering
        \includegraphics[width=\linewidth]{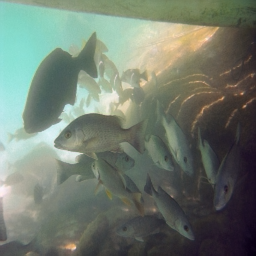} 
    \end{subfigure}%
    \hfill
    \begin{subfigure}{0.09\linewidth}
        \centering
        \includegraphics[width=\linewidth]{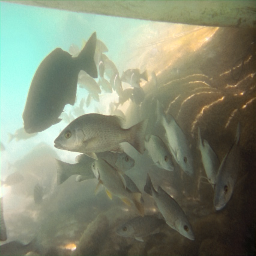} 
    \end{subfigure}%
    \hfill
    \begin{subfigure}{0.09\linewidth}
        \centering
        \includegraphics[width=\linewidth]{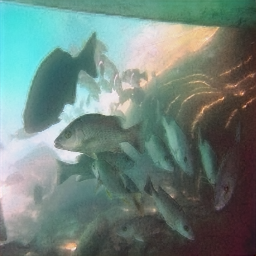} 
    \end{subfigure}%
    \hfill
    \begin{subfigure}{0.09\linewidth}
        \centering
        \includegraphics[width=\linewidth]{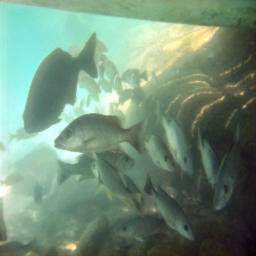} 
    \end{subfigure}%
    \hfill
    \begin{subfigure}{0.09\linewidth}
        \centering
        \includegraphics[width=\linewidth]{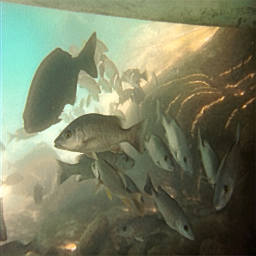} 
    \end{subfigure}%
    \vskip 0.5em

    \begin{subfigure}{0.09\linewidth}
        \centering
        \includegraphics[width=\linewidth]{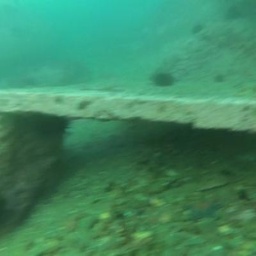} 
    \end{subfigure}%
    \hfill
    \begin{subfigure}{0.09\linewidth}
        \centering
        \includegraphics[width=\linewidth]{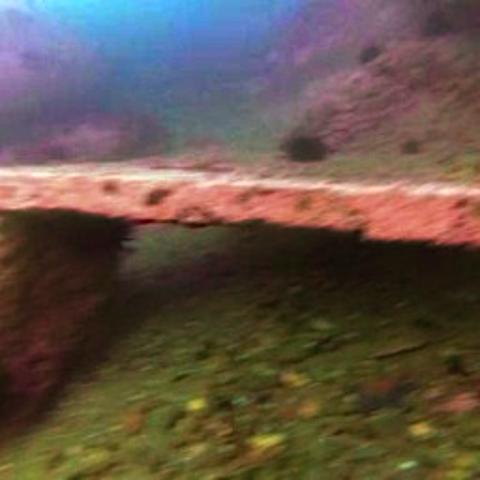} 
    \end{subfigure}%
    \hfill
    \begin{subfigure}{0.09\linewidth}
        \centering
        \includegraphics[width=\linewidth]{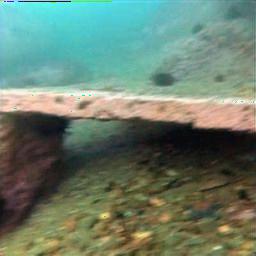} 
    \end{subfigure}%
    \hfill
    \begin{subfigure}{0.09\linewidth}
        \centering
        \includegraphics[width=\linewidth]{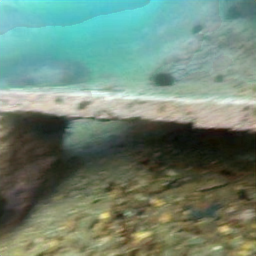} 
    \end{subfigure}%
    \hfill
    \begin{subfigure}{0.09\linewidth}
        \centering
        \includegraphics[width=\linewidth]{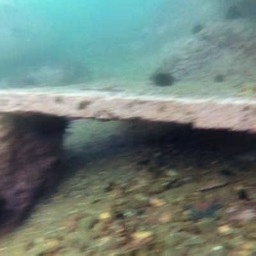} 
    \end{subfigure}%
    \hfill
    \begin{subfigure}{0.09\linewidth}
        \centering
        \includegraphics[width=\linewidth]{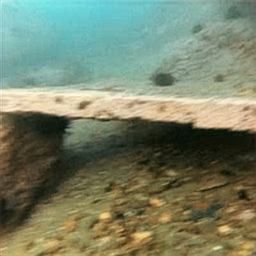} 
    \end{subfigure}%
    \hfill
    \begin{subfigure}{0.09\linewidth}
        \centering
        \includegraphics[width=\linewidth]{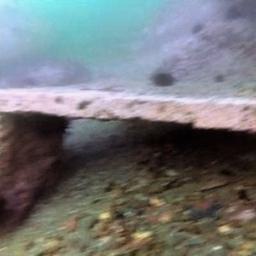} 
    \end{subfigure}%
    \hfill
    \begin{subfigure}{0.09\linewidth}
        \centering
        \includegraphics[width=\linewidth]{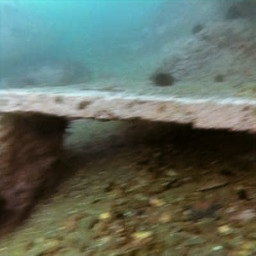} 
    \end{subfigure}%
    \hfill
    \begin{subfigure}{0.09\linewidth}
        \centering
        \includegraphics[width=\linewidth]{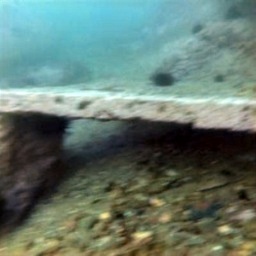} 
    \end{subfigure}%
    
    \caption{Visual comparisons on non-reference benchmarks UIEB-60 \cite{article12}, EUVP \cite{article44} and RUIE \cite{article65}. }
    \label{fig:five}
\end{figure*}

\subsection{Comparisons with the State-of-the-Art}

We compare the proposed Trans-UIE with seven state-of-the-art underwater image enhancement methods, including one conventional method, GDCP \cite{article72}, and six deep learning-based methods: WaterNet \cite{article12}, Ucolor \cite{article11}, PUIE-Net \cite{article33}, Semi-UIR \cite{article60}, U-shape \cite{article14}, and HUPE \cite{article73}. All competing methods were re-trained on our training dataset to ensure a fair comparison.

For experiments on reference datasets, \cref{tab:1} presents the quantitative evaluation results on the LSUI-400 and UIEB-90 datasets. On LSUI-400, our method achieves the best performance in terms of PSNR and SSIM, significantly outperforming competing methods. On UIEB-90, our method achieves the highest SSIM score, although its PSNR is slightly lower than Semi-UIR. This discrepancy may stem from the inclusion of the Pearson correlation loss during pre-training, which encourages the model to focus on the overall dataset distribution. Since the UIEB \cite{article12} dataset is smaller (800 pairs) compared to the larger LSUI \cite{article14} dataset (3879 pairs), the Pearson correlation loss causes the model to align more closely with the LSUI \cite{article14} dataset's distribution, resulting in lower scores on UIEB . A detailed discussion of how Pearson correlation loss mitigates overfitting risk is provided in subsequent ablation experiments. In addition to quantitative results, qualitative results are shown in \cref{fig:four}. Visually, our method provides pleasing results, accurately restoring underwater images and outperforming pseudo-label results. In contrast, competing methods exhibit issues such as color casts, noise, and over-enhancement.

For experiments on non-reference datasets, \cref{tab:2} presents the quantitative evaluation results on the UIEB-60 \cite{article12}, EUVP \cite{article44}, and RUIE \cite{article65} datasets. Our method significantly outperforms competing algorithms in terms of MUSIQ \cite{article61} metrics, while also achieving competitive results on UIQM \cite{article66} and UCIQE \cite{article67}. However, as noted in \cite{article74,article75,article12}, the UIQM \cite{article66} and UCIQE \cite{article67} metrics exhibit certain biases, which may not fully reflect the true visual quality of the restored images. \cref{fig:five} further illustrates the qualitative results on the non-reference datasets. Compared to existing methods, our approach robustly restores various underwater images with accurate colors and rich details, demonstrating excellent generalization across diverse underwater scenarios.

\subsection{Ablation Studies}

\noindent \textbf{Analysis of Transfer Learning Effectiveness.} We compared four methods: (a) \textbf{base}: the pre-trained model without fine-tuning; (b) \textbf{Trans-base}: fine-tuning with only the pixel loss \(L_{\text{pix}}\); (c) \textbf{Trans-base+\(L_{\text{vgg}}\)}: fine-tuning with pixel loss \(L_{\text{pix}}\) and perceptual loss \(L_{\text{vgg}}\); and (d) \textbf{Trans-UIE}: fine-tuning with all losses.

\begin{figure*}
    \centering
    \begin{subfigure}{0.18\linewidth}
        \centering
        \textbf{Input} 
    \end{subfigure}%
    \hfill
    \begin{subfigure}{0.18\linewidth}
        \centering
        \textbf{Base} 
    \end{subfigure}%
    \hfill
    \begin{subfigure}{0.18\linewidth}
        \centering
        \textbf{Trans-base} 
    \end{subfigure}%
    \hfill
    \begin{subfigure}{0.18\linewidth}
        \centering
        \textbf{Trans-base+\(L_{\text{vgg}}\) } 
    \end{subfigure}%
    \hfill
    \begin{subfigure}{0.18\linewidth}
        \centering
        \textbf{Trans-UIE} 
    \end{subfigure}%

    \vskip 0.5em
    
    \begin{subfigure}{0.18\linewidth}
        \centering
        \includegraphics[width=\linewidth]{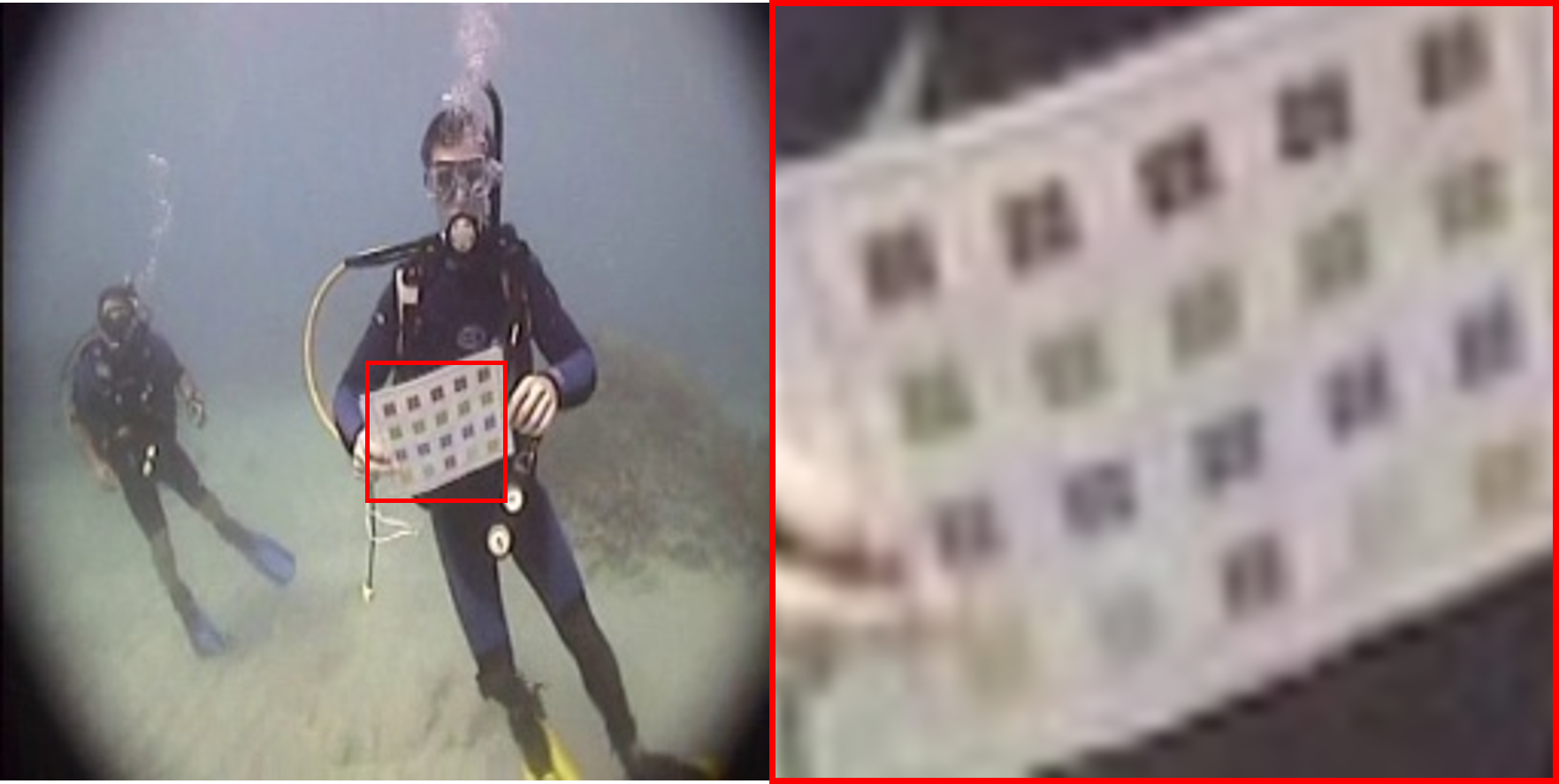} 
    \end{subfigure}%
    \hfill
    \begin{subfigure}{0.18\linewidth}
        \centering
        \includegraphics[width=\linewidth]{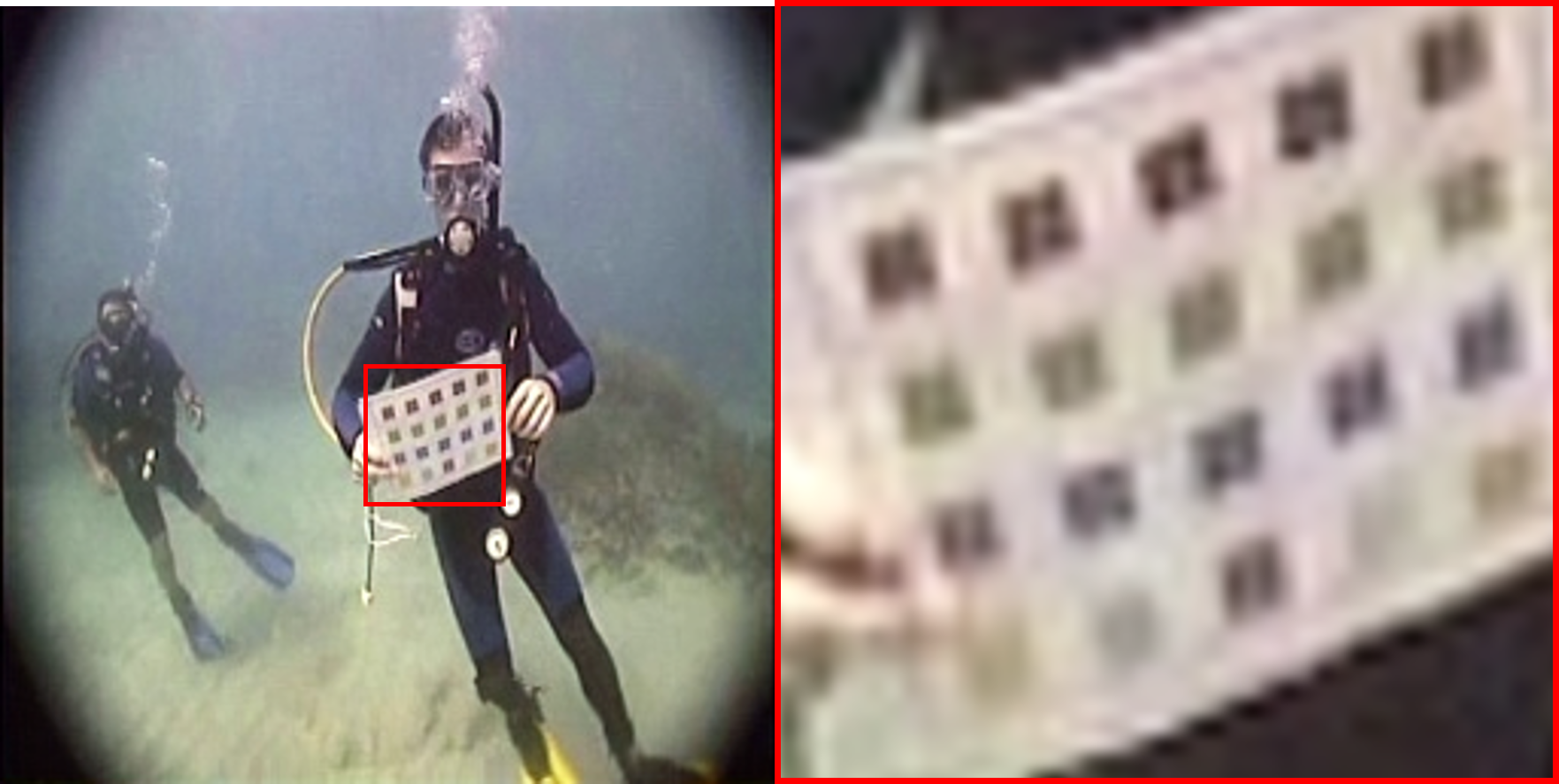} 
    \end{subfigure}%
    \hfill
    \begin{subfigure}{0.18\linewidth}
        \centering
        \includegraphics[width=\linewidth]{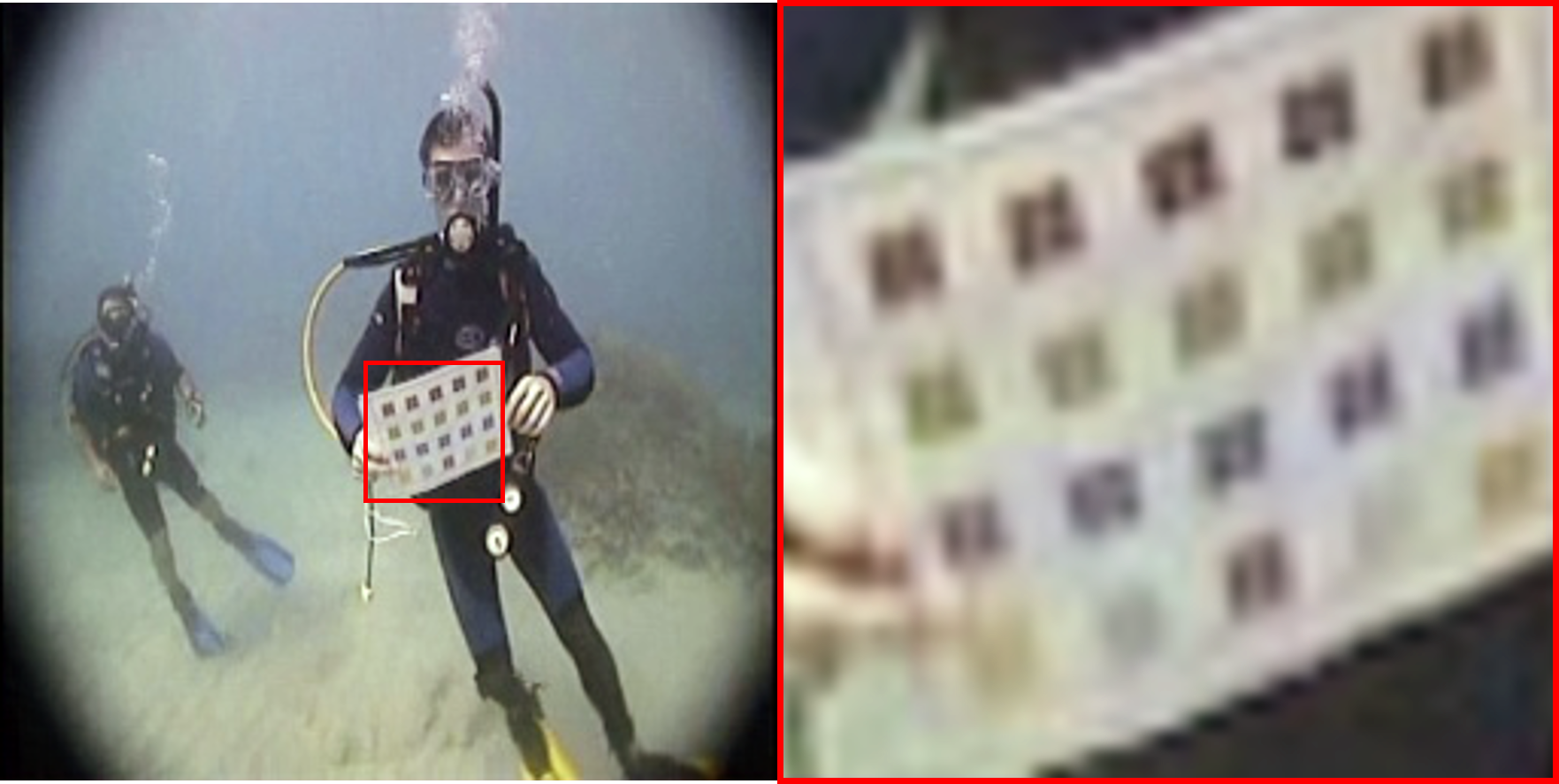} 
    \end{subfigure}%
    \hfill
    \begin{subfigure}{0.18\linewidth}
        \centering
        \includegraphics[width=\linewidth]{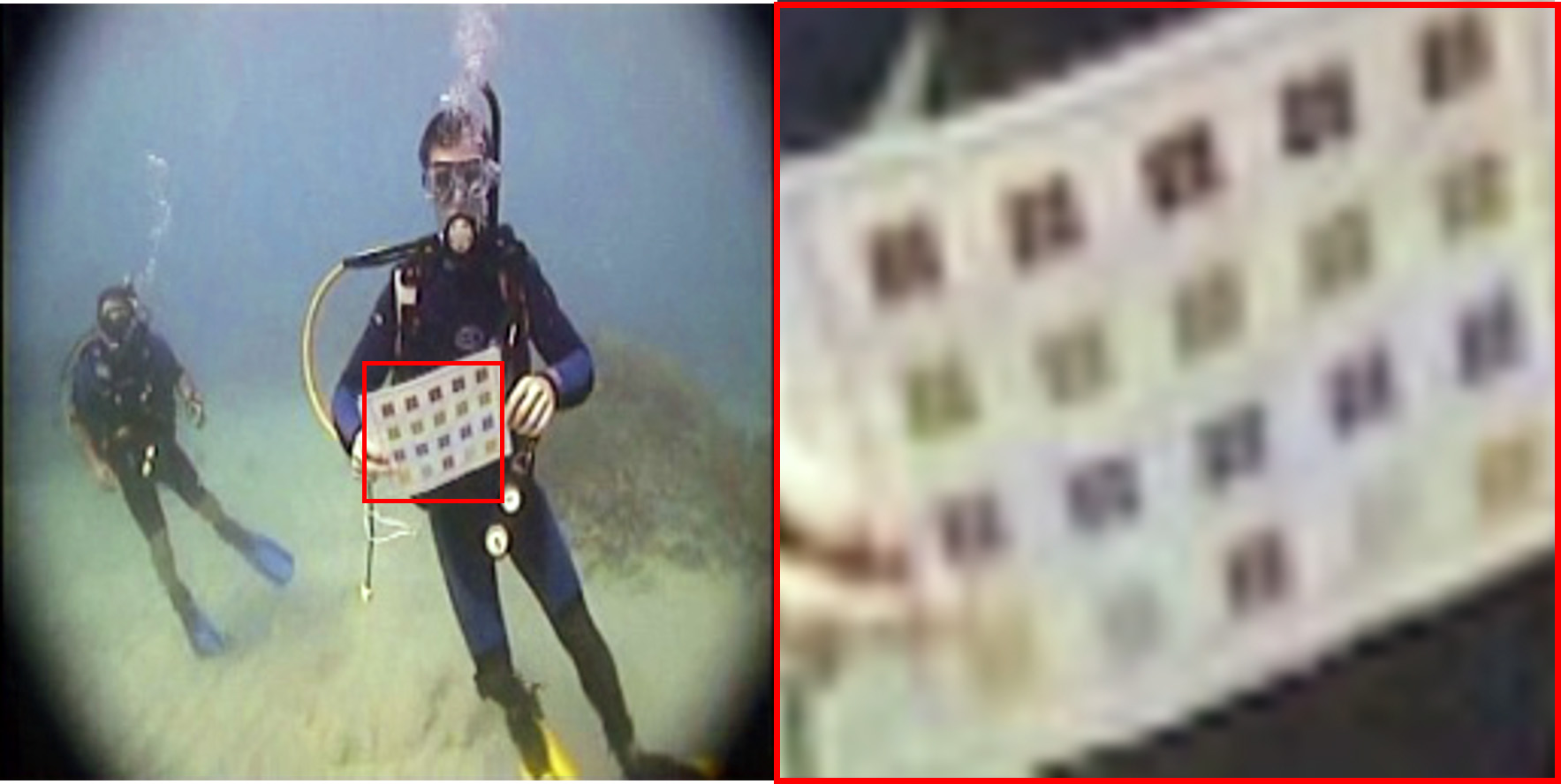} 
    \end{subfigure}%
    \hfill
    \begin{subfigure}{0.18\linewidth}
        \centering
        \includegraphics[width=\linewidth]{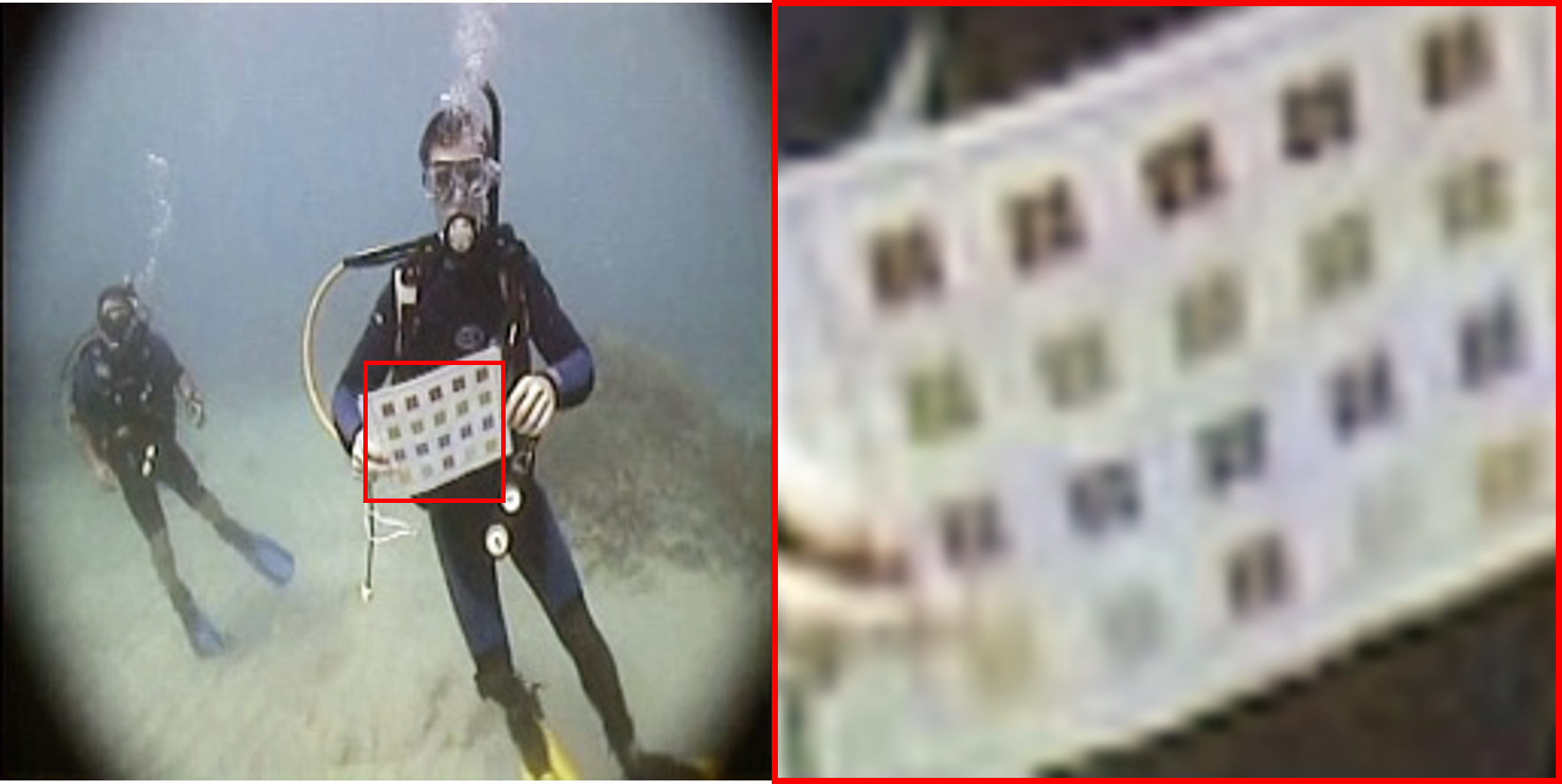} 
    \end{subfigure}%
    \caption{Results of Ablation Analysis on the Transfer Learning Effectiveness of Trans-UIE. }
    \label{fig:six}
    
\end{figure*}

Comparisons were conducted on both reference and non-reference datasets, with quantitative results presented in \cref{tab:3}. Our complete solution consistently outperforms the others. Comparing \textbf{base} and \textbf{Trans-base}, we observe that \textbf{Trans-base}, which uses pseudo-labels as training supervision, may amplify existing biases or errors, leading to confirmation bias. The introduction of perceptual loss \(L_{\text{vgg}}\) in \textbf{Trans-base+\(L_{\text{vgg}}\)} effectively mitigates this bias. Upon observing \textbf{Trans-UIE}, we note a decrease in PSNR scores on LSUI. This can be attributed to the introduction of evaluation loss \(L_{\text{MUSIQ}}\), which reduce the domain gap between underwater and real surface images. Qualitative results, shown in \cref{fig:six}, highlight that \textbf{Trans-UIE} delivers the best image restoration, with superior color fidelity and rich detail.

\begin{table}
\resizebox{\linewidth}{!}{
\begin{tabular}{c|ccccccc}
\hline
Method & Fine-Tune &  \(L_{\text{vgg}}\) & \(L_{\text{MUSIQ}}\) & LSUI-400 & EUVP & RUIE \\
\hline
\textbf{base}  &  &   &   & 30.51 & 48.59 & 35.68 \\
\textbf{Trans-base}  & $\checkmark$ &   &  & 27.56 & 45.81 & 34.77 \\
\textbf{Trans-base+\(L_{\text{vgg}}\)}  & $\checkmark$ & $\checkmark$ &   & 27.67 &  45.90 & 34.91 \\
\textbf{Trans-UIE}  & $\checkmark$ & $\checkmark$ & $\checkmark$ & 27.23 & 49.62 & 40.23 \\
\hline
\end{tabular}
}
\caption{ Ablation analysis of transfer learning effectiveness: tested on LSUI-400 \cite{article14}, EUVP \cite{article44}, and RUIE \cite{article65} benchmarks in terms of PSNR and MUSIQ. }
\label{tab:3}
\end{table}


\noindent \textbf{Analysis of Pearson Correlation Loss Effectiveness.} We compared two pre-trained models: (a) \textbf{base}: the pre-trained model using only pixel loss; (b) \textbf{UIR-Net}: the pre-trained model using both pixel loss and Pearson correlation loss, to assess the impact of incorporating Pearson correlation loss on model performance.

Comparisons were conducted on both reference and non-reference datasets, with quantitative results presented in \cref{tab:4}. The results show that our complete solution outperforms the others. \textbf{UIR-Net} exhibits decreased scores on the reference datasets, while improving performance on non-reference datasets. This can be attributed to the use of two reference datasets with significantly different sample sizes and a relatively small total dataset, which may lead to overfitting. The inclusion of Pearson correlation loss enables the model to focus on the linear relationships between data points, effectively mitigating overfitting and improving performance on non-reference datasets.

\begin{table}
  \centering
  \begin{tabular}{@{}lcccc@{}}
    \toprule
    Method & LSUI-400 & UIEB-90  & EUVP & RUIE \\
    \midrule
    \textbf{base} & 30.66 & 23.73 & 48.28 & 35.52 \\
    \textbf{UIR-Net} & 30.51 & 23.20 & 48.59 & 35.68 \\
    \bottomrule
  \end{tabular}
  \caption{Ablation analysis of Pearson correlation loss effectiveness: tested on LSUI-400 \cite{article14}, UIEB-90 \cite{article12}, EUVP \cite{article44}, and RUIE \cite{article65} benchmarks in terms of PSNR and MUSIQ. }
  \label{tab:4}
\end{table}


\noindent \textbf{Analysis of Feedforward Network Effectiveness.} We compared two pre-trained models: (a) \textbf{base}: the pre-trained model using a conventional feedforward network; and (b) \textbf{UIR-Net}: the pre-trained model employing a Channel Reordering Gated Feedforward Network (CRGFN). The comparison was conducted on reference datasets, with quantitative results presented in \cref{tab:5}.

The results demonstrate that our complete solution outperforms the others. CRGFN effectively enhances the representational capability of the pre-trained model, UIR-Net.

\begin{table}
  \centering
  \begin{tabular}{@{}lcc@{}}
    \toprule
    Method & LSUI-400 & UIEB-90 \\
    \midrule
    \textbf{base} & 28.73 & 21.66 \\
    \textbf{UIR-Net} & 30.51 & 23.20 \\
    \bottomrule
  \end{tabular}
  \caption{Ablation analysis of feedforward network effectiveness: tested on LSUI-400 \cite{article14} and UIEB-90 \cite{article12} benchmarks in terms of PSNR.}
  \label{tab:5}
\end{table}

\subsection{Evaluation Loss Hyperparameter Experiment}
\begin{figure}[htbp]
    \centering
    \begin{subfigure}{0.11\textwidth}
        \centering
        Base 
    \end{subfigure}%
    \hfill
    \begin{subfigure}{0.11\textwidth}
        \centering
        \( \alpha \) = 0.0001 
    \end{subfigure}%
    \hfill
    \begin{subfigure}{0.11\textwidth}
        \centering
        \( \alpha \) = 0.003 
    \end{subfigure}%
    \hfill
    \begin{subfigure}{0.11\textwidth}
        \centering
        \( \alpha \) = 0.1  
    \end{subfigure}%

    \vskip 0.5em
    
    \begin{subfigure}{0.11\textwidth}
        \centering
        \includegraphics[width=\linewidth]{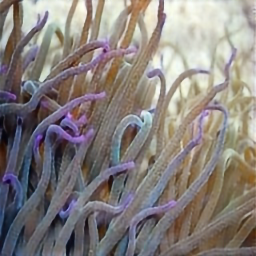} 
    \end{subfigure}%
    \hfill
    \begin{subfigure}{0.11\textwidth}
        \centering
        \includegraphics[width=\linewidth]{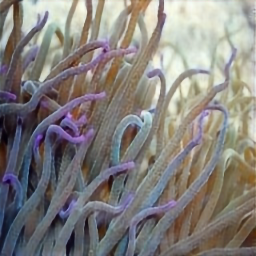} 
    \end{subfigure}%
    \hfill
    \begin{subfigure}{0.11\textwidth}
        \centering
        \includegraphics[width=\linewidth]{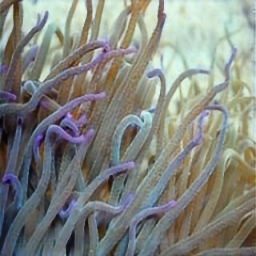} 
    \end{subfigure}%
    \hfill
    \begin{subfigure}{0.11\textwidth}
        \centering
        \includegraphics[width=\linewidth]{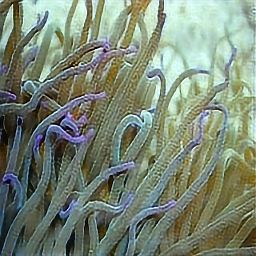} 
    \end{subfigure}%
    
    \caption{Examples based on different NR-IQA score loss hyperparameter.}
    \label{fig:seven}
\end{figure}

To assess the impact of different values for the NR-IQA score loss hyperparameter \(\alpha\), we compared the following settings: (a) \textbf{base}: processed only by the pre-trained model; (b) \(\alpha = 0.0001\); (c) \(\alpha = 0.003\); (d) \(\alpha = 0.1\). Results are shown in the \cref{fig:seven}.

From the comparison, it is evident that when \(\alpha\) is too large, the model overemphasizes the distribution of real-world images, leading to color bias. Conversely, when \(\alpha\) is too small, the effect is negligible. Thus, we selected \(\alpha = 0.003\) as the optimal value.

\subsection{Breakdown of the Training}
\begin{figure}[htbp]
    \centering
    \begin{subfigure}{0.11\textwidth}
        \centering
        Input 
    \end{subfigure}%
    \hfill
    \begin{subfigure}{0.11\textwidth}
        \centering
        Batches 200 
    \end{subfigure}%
    \hfill
    \begin{subfigure}{0.11\textwidth}
        \centering
        Batches 600 
    \end{subfigure}%
    \hfill
    \begin{subfigure}{0.11\textwidth}
        \centering
        Batches 1000  
    \end{subfigure}%
    \vskip 0.5em
    \begin{subfigure}{0.11\textwidth}
        \centering
        \includegraphics[width=\linewidth]{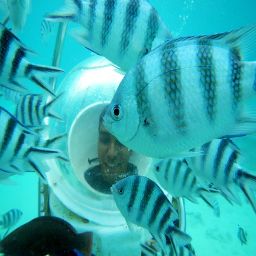} 
    \end{subfigure}%
    \hfill
    \begin{subfigure}{0.11\textwidth}
        \centering
        \includegraphics[width=\linewidth]{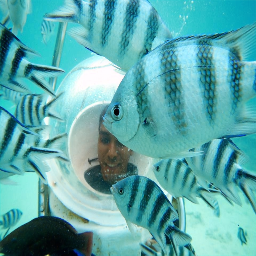} 
    \end{subfigure}%
    \hfill
    \begin{subfigure}{0.11\textwidth}
        \centering
        \includegraphics[width=\linewidth]{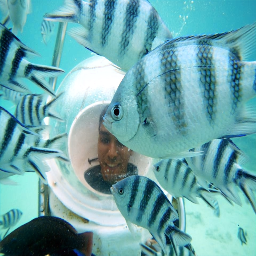} 
    \end{subfigure}%
    \hfill
    \begin{subfigure}{0.11\textwidth}
        \centering
        \includegraphics[width=\linewidth]{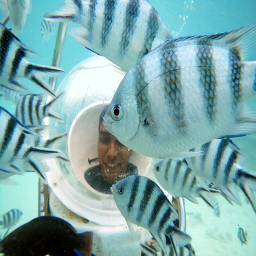} 
    \end{subfigure}%
    
    \caption{Examples of intermediate predictions. }
    \label{fig:eight}
\end{figure}

To further illustrate the transfer learning process, we present sample images from the training procedure. The results are shown in \cref{fig:eight}. From the comparison, it is evident that as training progresses, the image quality progressively improves.

\section{Conclusion}
We propose an efficient transfer learning-based underwater image enhancement method named Trans-UIE. As demonstrated by the ablation experiments, the proposed method outperforms other SOTA algorithms, which can be attributed to its effective reduction of feature distribution shift and domain discrepancy. Future research can be directed in two areas: (1) expanding the transfer learning framework to cover other restoration tasks, and (2) enhancing performance through improved memory management.

{
    \small
    \bibliographystyle{ieeenat_fullname}
    \bibliography{main}
}

\end{document}